\documentclass{IOS-Book-Article}

\usepackage{mathptmx}
\usepackage{graphicx}
\usepackage{titlesec}
\usepackage{url}
\usepackage{multirow}
\usepackage{blindtext}
\usepackage{graphicx}
\graphicspath{{images/}}

\def\hb{\hbox to 10.7 cm{}}

\begin{document}

\pagestyle{headings}
\def\thepage{}

\begin{frontmatter}              % The preamble begins here.

%\pretitle{Pretitle}
\title{Aspect Classification for Legal Depositions}

\markboth{}{September, 2020\hb}
%\subtitle{Subtitle}

\author{Saurabh Chakravarty%
\thanks{Corresponding Author: Saurabh Chakravarty, Department of Computer Science, Virginia Tech, Blacksburg, VA 24061, USA; Email:saurabc@vt.edu}},
\author{Satvik Chekuri},
\author{Maanav Mehrotra},
\author{Edward A. Fox}
\address{Virginia Tech, Blacksburg, VA 24061 USA}
% \author[B]{\fnms{Second} \snm{Author}}
% and
% \author[B]{\fnms{Third} \snm{Author}}

% \runningauthor{B.P. Manager et al.}
% \address[A]{Book Department, IOS Press, The Netherlands}
% \address[B]{Short Affiliation of Second Author and Third Author}

% \address[B]{Short Affiliation of Second Author and Third Author}

\begin{abstract}
Attorneys and others have a strong interest in having a digital library with suitable services (e.g., summarizing, searching, and browsing) to help them work with large corpora of legal depositions. Their needs often involve understanding the semantics of such documents. That depends in part on the role of the deponent, e.g., plaintiff, defendant, law enforcement personnel, expert, etc. In the case of tort litigation associated with property and casualty insurance claims, such as relating to an injury, it is important to know not only about liability, but also about events, accidents, physical conditions, and treatments.

We hypothesize that a legal deposition consists of various aspects that are discussed as part of the deponent testimony. Accordingly, we developed an ontology of aspects in a legal deposition for accident and injury cases. Using that, we have developed a classifier that can identify portions of text for each of the aspects of interest. Doing so was complicated by the peculiarities of this genre, e.g., that deposition transcripts generally consist of data in the form of question-answer (QA) pairs. Accordingly, our automated system starts with pre-processing, and then transforms the QA pairs into a canonical form made up of declarative sentences. Classifying the declarative sentences that are generated, according to the aspect, can then help with downstream tasks such as summarization, segmentation, question-answering, and information retrieval.

Our methods have achieved a classification F1 score of 0.83. Having the aspects classified with a good accuracy will help in choosing QA pairs that can be used as candidate summary sentences, and to generate an informative summary for legal professionals or insurance claim agents. Our methodology could be extended to legal depositions of other kinds, and to aid services like searching.
\end{abstract}

\end{frontmatter}
\markboth{September, 2020\hb}{September, 2020\hb}

%\thispagestyle{empty}
%\pagestyle{empty}

% \maketitle

\section{Introduction}
Legal deposition documents are comprised of dialogue exchanges between attorneys and deponents, mainly focused on identifying observations and the facts of a case.
These conversations are in the form of (possibly quickfire) question-answer (QA) sets. 
As with some other discussions, these QA pairs may only loosely observe linguistic guidelines.
Sometimes the resulting sentences (or sentence fragments) are not well-formed. 
Often an answer cannot be understood independently of the prior question or may depend on a question posed even earlier.

Processing and comprehension of these types of documents by humans are difficult, time-consuming, and leads to considerable expense since the work is typically carried out by attorneys and paralegals.
This process often adds to the elapsed time when handling a case or preparing for trial and causes real-time staffing problems before trials.
Automatic processing and comprehension of the content present in legal depositions would be immensely useful for law professionals, helping them to identify and disseminate salient information in key documents, as well as reducing time and cost.

Processing deposition documents using traditional text processing techniques is a challenge because of the syntactic, semantic, and discourse characteristics of the QA conversations.
These make it difficult to apply conventional natural language processing (NLP) methods, including co-reference resolution and summarization techniques.
Identifying the key concepts using NLP based rules is another challenge.
%In many corpora, the root words that are most prevalent in sentences help identify the core concepts present in a document. 
Often, the form of deposition documents does not facilitate capturing core concepts using sentence-oriented methods, since the context of the conversation is spread across the questions and answers.

Work such as \cite{chakravarty_jurix} proposed methods to transform the QA pairs into a canonical form made up of declarative sentences.
Such conversion allows the text to be processed for downstream processing.
One such task is summarization.
However, preliminary experiments starting with the declarative sentences resulting from the transformation, and feeding them into a summarization method pre-trained on news article corpora, led to poor results.
%However, for tasks such as summarization, using the declarative sentences as text to generate summaries of legal depositions results in summaries that are poor. 
The summaries thus generated omit important information and lack the cohesion and context required to be comprehended and used by a legal professional.
%While there are a few parts that are coherent and present with context, the summary is not informative.
Partly, this happens because of a lack of domain understanding on the part of these methods, which
%Most of these methods 
have been trained on news articles that are markedly different in content and structure from legal depositions.
For example, the news covers popular topics, subjects, people, organizations, and/or locations, as distinct from the specialized matters in legal disputes.
Further, key concepts often can be identified in the news based on repetition, or based on formulaic coverage of ``who, what, when, where, why, how'' discussions, while depositions often provide such coverage in background warm-up sections that attorneys largely ignore later.
Regarding structure, news stories have the most significant concepts mentioned in the beginning.
Then these concepts are discussed in more detail further along in the article. 
%Summarization systems identify words and concepts that are repeated across the document. 
Accordingly, the sentences with these words and concepts are identified as summary candidates that are processed further to generate either an extractive or abstractive summary. 
%Legal depositions do not follow such a structure. 
On the other hand, many important parts of depositions are present in the middle or end segments in the document. 
There may be a very little repetition of some key concepts, such as when a key admission is made, and an attorney deliberately avoids allowing such a statement to be elaborated upon.
In such cases, facts, once collected as part of the deposition, are rarely repeated in any other part of the deposition. This non-repetition thus diminishes the utility of simple frequency-based scores, that work so well in the news domain.

For a legal deposition summary, we want the system to pick the content from the document in an objective and unbiased way.
It is also important that the information included in the summary is factually correct.
To comprehend a legal deposition better, understanding the requirements of the downstream tasks from a domain perspective would be useful.
Consumers of legal deposition summaries are more interested in important parts that relate to the case pleadings or claim complaints.
It is more important to have coverage of the core events that are mentioned in the pleadings and complaints as opposed to uniform coverage of the whole deposition document.
Coverage of the details of the event and the events before and after it, often are important for legal professionals.

Another challenge with legal depositions is that there are different deponent roles pertaining to a case.
The focus of questioning varies based on the deponent type.
Accordingly, different deponent types lead to different mixtures of aspects being discussed during their depositions.
Grouping the aspects based on the deponent type would allow us to structure a summary better in terms of aspect distribution and layout.
% We hypothesize that breaking a deposition into its different topics that cover the various aspects of interest would lead to better summaries.

Events, entity mentions, and facts are present throughout the length of the deposition.
Identifying the aspects covered in a deposition would allow a deposition to be broken up into its constituent topical parts.
Summaries can be generated based on a predefined distribution and layout of different aspect sentences present in the deposition.
Such a layout and distribution can be learned from existing legal depositions and summaries, and could be further refined based on case pleadings and deponent types.
Identifying the various aspects of interest present in a legal deposition would help in other downstream tasks in addition to generating summaries.

The aspects in a deposition can also help in focused information retrieval, where aspects are associated with facets used in searching and browsing.
%Events, entity mentions, and facts are present throughout the length of the deposition.
Thus, a deposition could be presented to the end-user using aspects for one class of facets, letting them gauge the deposition content and aspect distribution at a high-level and explore further by digging into each facet or combination of facets.
An example of such exploration by a legal professional would be a review of all the segments that are of the aspect of type \textit{``Event Details"}, to ascertain the specific facts related to the actual event (or accident).
Performing a focused search based on aspects would help retrieve results that are more relevant based on the chosen aspect.
Consider, for example, the role of a witness just before an accident. 
A search of person entity mentions for the relevant aspect of \textit{``Event Background"} would present the user with the relevant results.
Another important use of aspects would be to identify segments within a legal deposition that have the same thematic context and topical coherence.
The segment metadata could be assigned an aspect label based on the majority aspect within the segment.
These segments, with their associated aspect metadata, could be used to speed up the exploration of a set of legal depositions pertaining to a case.
%An example of such exploration by a legal professional would be a review of all the segments that are of the aspect of type \textit{"Event Details"}, to ascertain the specific facts related to the actual event (or accident).
%A legal deposition can also be presented to the end-user using aspects for one class of facets, letting them gauge the deposition content and aspect distribution at a high-level and explore further by digging into each facet or combination of facets.

This work focuses on classifying the various aspects of depositions into a predefined ontology that pertains to the legal domain.
As part of our initial work, we analyzed legal depositions and their summaries generated by legal professionals.
These were legal depositions for different cases and varying deponent roles.
We started with an ontology of 20 aspects that were later trimmed down to a set of 12 by merging aspects that were similar in nature.
The ontology is described in detail in Section \ref{sec:Methods-Ontology}.

The core contributions of our work are as follows:
\begin{enumerate}
    \item An ontology of aspects for accident and injury cases that can be expanded/modified to other kinds of cases.
    \item A grouping of relevant aspects based on deponent types that can be expanded/modified for other kinds of cases.
    \item Classification methods to identify the various aspects of QA pairs in the deposition.
    \item A dataset that can be used by the community to use and expand their research in legal and other domains.
    \item An overall framework to identify the aspects of legal depositions that can be re-purposed for other kinds of cases in the legal domain. This framework also could be used for other domains, like scientific and medical literature summarization, etc.
\end{enumerate}

This paper is organized as follows.
Section \ref{sec:RelatedWork} discusses the related work in the field of question-answer corpora processing and classification.
Section \ref{sec:Methods} presents the methodology for the classification of aspects.
In Sections \ref{sec:Dataset} and \ref{sec:Experiments}, the dataset and the experiment results are reported.
Then the conclusion and plans for future work are presented in Section \ref{sec:Conclusion}.

\section{Related Work}
\label{sec:RelatedWork}

As part of our literature search, we studied works that relate to the comprehension of conversations.
Work in \cite{DA} introduced the concept of dialog acts (DA) in spoken conversations.
A DA represents the communicative intent behind the speaker's intent in a two or multi-party conversation.
Work in \cite{da_classic1,da_classic2,da_classic3,da_classic4,da_classic5,da_classic6,da_classic7,da_classic8} used different techniques to classify the dialog acts in different settings such as text chats, meetings, etc. 
Identifying the conversation sentences into DAs provides a way to understand the discourse structure of the conversation.
% The text can be processed further for other downstream tasks based on the DAs.
% One such use is to transform a question-answer (QA) pair into a canonical or declarative form.
A challenge with spoken conversation sentences is that the context of the conversation is spread across the question and the answer.
This is especially pronounced in the area of legal depositions.
One method to process the QA sentences in an efficient way is to fuse the question and answer together into a single or series of sentences.
Work in \cite{chakravarty_jurix} used the DAs of the questions and answers to transform a QA pair to a declarative sentence.
The QA pairs were from a corpus of legal depositions \cite{tobacco-link}.
Transformation rules were developed to break the question and answer sentences into chunks, and words were permuted, deleted, and added from the question and answer sentences, followed by a fusion of the modified question and answer sentences.
Different transformation rules were developed based on the combination of the question and answer DA classes.
The authors used the DA ontology for legal depositions from the work in \cite{da_for_qa} to frame the rules for transformation.
Transforming the QA pairs to a text representation that is conducive to other downstream tasks would be useful; here we explore further whether a transformation to a canonical form will help achieve better results.

% % Justify and flow into the summarization problem and its related problems. And how aspect classification can help.
% Summarization methods in \cite{summ_classical_dorr, Summ_Rush2015, Summ_See2017, Summ_Liu2019, summ_paulus2017deep, Summ_Nallapati2016} use a combination of statistical and neural techniques to generate extractive and abstractive summaries.
% Work in \cite{socher_summ_critique} highlights the limitations of the current summarization methods landscape.
% It argued that, though the recent works have established the state of the art results in the summarization datasets like CNN/DailyMail \cite{Summ_Nallapati2016}, they only slightly outperform the lead-3 baseline.
% The layout bias that is present in these systems does not make them generalized so that they could be used to generate summaries of documents from a different domain such as legal, scientific. and medicine.
% In order to generate informative summaries for legal depositions, analyzing existing summaries that exist in the domain would be useful.
% The method to generate a summary for legal deposition is very different than news articles.
% Attorneys and paralegals want to zero in on the actual case facts and various events that relate to the case.
% These segments in a legal deposition are hard to catch since the traditional summarization algorithms are trained in a different way.

Classical classification methods such as Naïve Bayes \cite{classical_NB}, decision tree \cite{classical_DT}, k-nearest neighbor \cite{classical_nearest_neighbor}, association rules \cite{classical_association_rules}, etc. have been used for classifying text.
These methods have been combined with various feature selection techniques such as gini index~\cite{fs_gini}, conditional entropy \cite{fs_conditional_entropy}, ${\rm \chi^{2}}$-statistic \cite{fs_chi_square}, and mutual information \cite{fs_pmi}.
They have also been augmented with various feature engineering techniques to improve the classification further.

Another class of classification methods uses rich word embeddings like GloVe \cite{glove} and word2Vec \cite{word2vec}.
With them, good results have been reported in various text classification tasks.
Using the embeddings helps create a feature representation of the text without the need to perform complicated feature selection or engineering.
Methods such as \cite{iyyer2015deep, castro2017classifying} used the averaging of the embeddings of the sentence words. 
Though simple, these methods were able to perform competitively in many tasks.
However, a challenge with the classical feature selection and the word embedding based methods is that the created representation is a bag-of-words representation and does not factor in the ordering of the words.

Methods based on Deep Learning have further improved upon the established state-of-the-art results in text classification.
Sophisticated architectures like encoder-decoder models \cite{encoder_decoder_cho2014properties} along with attention \cite{attention_bahdanau2014neural} were equally effective in text classification.
These architectures are very deep and sometimes have multiple layers and a large number of parameters.
However, achieving this high-performance requires training of these models on a large dataset.

Another recent addition in the text representation space is the introduction of systems that can generate embeddings for sentences.
The premise of these systems was to learn a language model (LM) using a large corpus.
These models were trained using the principle of self-supervision on targeted NLP tasks like missing word prediction and next sentence classification.
As the model is trained on the targeted task, a side benefit of such training was the intermediate representation that is learned by the system to represent sentences.
These sentence embeddings are semantically very rich.

Work on BERT \cite{devlin2018bert} trained a language model  using a large corpus of English text.
The core contribution of this work was the generation of pre-trained sentence embeddings that have been learned using the left and right context of each token in the sentence. 
The system was trained in a bi-directional way to learn the semantic and syntactic dependencies between words in both directions.
After adding a fully connected neural network layer to the pre-trained embeddings, the authors propose that these embeddings can be utilized to model any custom NLP task.
It eliminates the need to create a custom network architecture.  
BERT internally uses the ``Transformer'' or the multi-layer network architecture presented in \cite{vaswani2017attention} to model the input text and the output embedding.
The trained system was able to outperform the established state-of-the-art in 11 benchmark NLP tasks.

As part of this paper, we explore the use of deep learning based methods and BERT sentence embeddings for our classification experiments.

\section{Methods}
\label{sec:Methods}

As part of our methods, we defined an ontology of aspects for the legal domain.
We also developed classification methods to classify the aspects for the QA pairs in the dataset, as per the defined ontology.
The following sections describe the ontology and the classification methods in more detail.

\subsection{Aspects Ontology}
\label{sec:Methods-Ontology}
% ToDo: Write how the aspects were identified via going through depositions and summaries. Paragraphs were observed in summaries.
% ToDo: Different deponent types had different aspects.
% ToDo: Talk about consultation with legal professionals about their opinion.
To identify the aspects, we analyzed the legal depositions and their summaries in our dataset.
The summaries in the dataset were arranged in paragraphs, where each aspect in the summary was present in a different paragraph.
However, there were multiple paragraphs that covered the same aspect.
Two authors of our paper created their own list of aspects initially. 
They collated all of the aspects together based on their mutual agreement.
Aspects were also grouped as per the deponent type.
These aspects were then reviewed by a legal professional and further refined to a final list of 12 aspects.
% The refinement of aspects is performed after carefully considering the training data distribution after getting annotated to the respective aspects. Not much of the data was available for a few of the aspect classes, sometimes even returning zero sentences. Also, an overlap of topics/context was existing between a few aspects leading to mis-classifications among these classes. Keeping these points in mind, and after consulting legal professionals, we have consolidated a few aspects. For example, Injury Description/Diagnosis and Treatment Received majorly cover the plaintiff's medical condition, hence the former aspect is merged into the TR aspect. 

Table ~\ref{tab:aspects_list} lists all the aspects with their descriptions.
Table ~\ref{table:deponent_roles} lists the deponent roles, their definition, and the aspects related to the respective roles.

\begin{table*}[htbp]
\centering
  \caption{List of aspects for accident/claim cases}
    \begin{tabular}{|p{0.06\linewidth}|p{0.20\linewidth}|p{0.68\linewidth}|}
\hline
\textbf{Topic   ID} & \textbf{Topic}                                                     & \textbf{Definition}                                                                                                                                                                                                                                               \\ \hline
1 & B (Biographical) & This topic covers the background of the witness, family and work history, along with educational background, training, etc.\\ \hline
2  & EB (Event   Background)  & This topic covers events that happened or conditions that existed just before the actual event (accident) that resulted in the legal claim.                                                                                                                 \\ \hline
3                   & ED (Event   Details)                                               & This topic covers all details about the accident event that resulted in the legal claim.                             \\ \hline
4                   & EC (Event   Consequences)                                          & This   topic covers the results or effects of the event that resulted in the legal claim, including injuries, pain, medical treatment, lost income, and impact of the injury/accident on the person's life.                                                   \\ \hline
5                   & PPC (Prior   Physical Condition)                                   & This topic covers what the injured person could do before this injury happened.                                                                                                                                                                                 \\ \hline
6                  & TR (Treatments   Received)                                         & This topic covers all medical treatment received by the plaintiff for the injury.   It includes EMT services, diagnostic testing, hospitalization, medications,   surgeries, medical appliances, therapy, and counseling.                                      \\ \hline
7                   & EE (Expert   Elaboration)                                          & This topic covers any detailed explanation by an expert witness. It usually involves the use of precise medical, engineering, vocational or economic terminology, and may include detailed elaboration on the definition of the term.                     \\ \hline
8                   & IP (Impact on Plaintiff)                                         & This topic covers any description of the physical, mental, emotional, or financial impact of the injury on the plaintiff, including physical limitations, recovery progress, and any planned or potential future treatment.                                  \\ \hline
9                  & DP (Deposition   Procedures)                                       & This topic covers the instructions that are often provided to deponents.                                                                                                                                                                                        \\ \hline
10                  & OPS (Operational procedures/ inspections   / maintenance / repairs) & Most   injury claims involve movable (cars, boats, etc.) or immovable property   (buildings, equipment, etc.). This topic covers the condition, operational procedures, inspection, maintenance, or repairs of the property involved in the accident/event. \\ \hline
11                  & PRD (Plaintiff-related   Details)                                  & For fact witnesses other than the plaintiff, this topic covers information gathered from them about the plaintiff.                                                                                                                                            \\ \hline
12                  & O (Other)                                                          & This is to be used for any topic that the annotator believes is not covered in the list above.                                                                                                                                                                \\ \hline
\end{tabular}

  \label{tab:aspects_list}
\end{table*}

\begin{table*}[h!]
\caption{Deponent Roles}
\centering
\begin{tabular}{|p{.12\textwidth}| p{.68\textwidth}| p{.14\textwidth}|}
 \hline
 \textbf{Deponent Role} & \textbf{Definition} & \textbf{Related Aspects} \\
  [0.5ex]
 \hline
 Plaintiff & This is generally the person who files the case against the defendant requesting damages for injury caused by the defendant in an event. & B, EB, ED, EC, PPC, TR, IP\\ \hline
 Fact Witness & This person is either a witness to the event or knows facts about when/where/how the accident/injury occurred. There is sometimes an interaction involved with the plaintiff, but it might not be in all the cases. & B, EB, ED, EC, PRD\\ \hline
 Expert Witness & This witness is brought in for their domain expertise pertaining to a case. These include medical, engineering, and other domain experts typically. & B, PPC, TR, EE, IP\\ \hline
 Related Organization Witness & These witnesses are from an organization that also is involved in the case. These could range from the defendant parties to organizations that were involved in the claim. & B, ED, IP, OPS, PRD\\ \hline
 Defendant & This is the party that has been sued. It could be a person who is being sued, or this could be representative of the organization being sued. A witness from an organization has some background related to the event. This background could range from inspection, maintenance, and upkeep of the real or personal property, to knowledge about past events that may be related to the event. & B, EB, ED\\ \hline
\end{tabular}

\label{table:deponent_roles}
\end{table*}

\subsection{Aspect Classification}

For our work, we used 3 different methods for classification. 
Two of the methods used deep neural networks to model the sentence embeddings for the input sentences. 
The third method used the BERT \cite{devlin2018bert} model to generate sentence embeddings. 
We wanted to measure whether simple architectures based on a rich sentence embedding would perform competitively compared to a deep neural network.
The following sections describe the architecture of these methods.

\subsubsection{CNN based classifier}
\begin{figure}[h!]
    \centering{
    \includegraphics[width=0.5\textwidth]{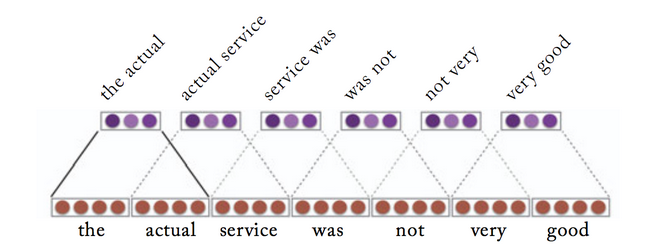}
    \caption{An n-gram convolution filter \cite{cnn_kimconvolutional,goldberg2017neural}.}
    \label{fig:CNN-Ngram}
}
\end{figure}

Though word embedding based deep averaging methods \cite{iyyer2015deep} were able to achieve competitive performance in text classification, one of the challenges with such methods is that they do not account for the word order in the sentence.
The final averaged representation of the sentence is the same, as long as it has the same words.
This follows a simple analogy with a bag-of-words model.
Work in \cite{cnn_kimconvolutional} used a convolutional neural network (CNN) to create a classifier that would be able to account for the word order in the sentence that has to be classified.
The network used a spanning window of a variable size, which was generally 2-4, to represent the n-gram based representation in a sentence.
The convolution filter size of the network was parameterized and was used to control how many n-grams will be run through the convolution process.
Figure \ref{fig:CNN-Ngram} depicts the convolution operation capturing a bi-gram representation. 

\begin{figure}[h!]
    \centering{
    \includegraphics[width=0.5\textwidth]{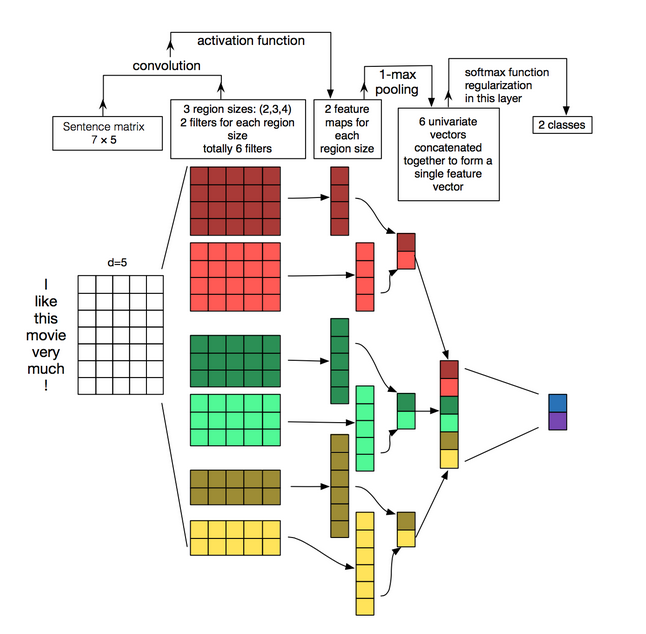}
    \caption{CNN based classifier architecture \cite{cnn_kimconvolutional,cnn_understanding}.}
    \label{fig:CNN-arch}
}
\end{figure}

We used an architecture similar to \cite{cnn_kimconvolutional} for our experiments.
The input sentence was tokenized into words.
The words were transformed into their embeddings using word2vec.
The convolution and max-pooling operations convert the word embeddings of the input words into a fixed-sized representation. 
This fixed-size representation is the sentence embedding that is generated by the network. 
Such an embedding, learned after the training process, is semantically rich since it captures the semantic and syntactic relationships between the words.
This sentence representation was used by a following fully connected layer, followed by a softmax output layer for classification. 
Figure \ref{fig:CNN-arch} shows the architecture of the whole network.

\subsubsection{LSTM with attention classifier}
\begin{figure}[h!]
    \centering{
    \includegraphics[width=0.5\textwidth]{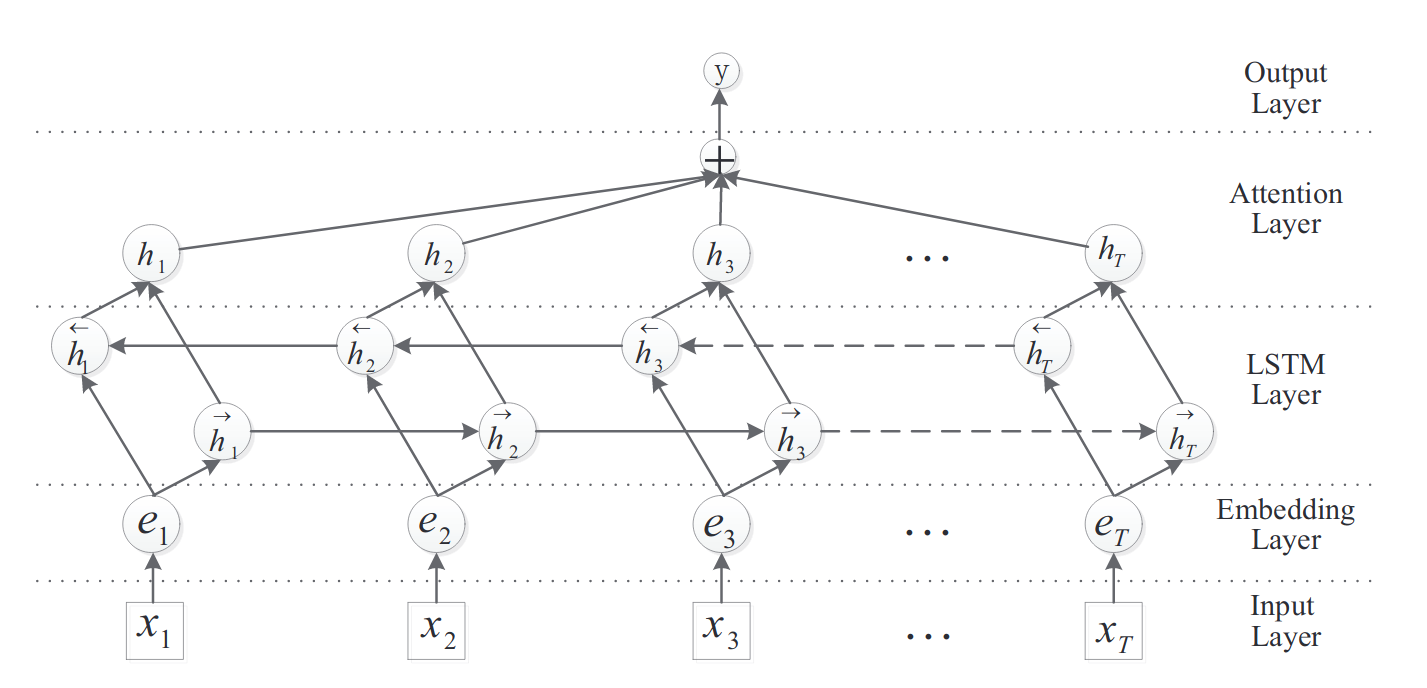}
    \caption{Bi-directional LSTM with attention architecture \cite{bi-lstm-attention}.}
    \label{fig:bi-lstm}
}
\end{figure}

Even though a CNN based network can capture the semantic and syntactic relationships between the words using the window based convolution operation, it struggles to learn the long term dependencies occurring in longer sentences.
Recurrent networks like LSTM \cite{lstm} have the ability to learn such long term dependencies in long sentences (which are often found in depositions).
Work in \cite{bi-lstm-attention} used a bi-directional LSTM with an attention mechanism for neural machine translation (NMT). 
The context of the input words makes their way back into the beginning of the recurrent network during the back-propagation step in the training phase.
This enables the system to learn long-term dependencies.
Also, since the network was bi-directional, it also learned the relationship of a word with the words that preceded it, in addition to the words that follow.
The hidden states of the network were joined to an attention layer that assigns a weight, known as attention weight, to each term. 
These attention weights capture the relative importance of the words based on the task. 
Though the work was used for NMT, we used the same network for classification by joining the attention layer to a dense layer followed by a softmax classification layer.
The system was trained end-to-end on the data, and the word embeddings were also learned as part of the training.
We added dropout \cite{dropout} to the embedding, LSTM, and penultimate layers.
Additionally, the L2-norm based penalty was applied as part of the regularization.
Figure \ref{fig:bi-lstm} shows the architecture of the network.

\subsubsection{BERT embeddings based classifier}
Work in \cite{vaswani2017attention} introduced a new approach to sequence to sequence architectures. 
In the recent past, RNNs and LSTMs have been used to model text and capture their long term dependencies. 
The challenge with the encoder-decoder model is that the output of the encoder is still a vector. 
% A sentence cannot be simple encoded in a vector.
A sentence loses some of its meaning when it is converted into a single vector via recurrent connections.
This loss is even greater when the sentence is long. 
The architecture proposed in the work did not have any recurrent connections to model the input.
The approach used stacked multiple layers of attention in the encoder.
The decoder also used the same architecture, but employed a masked multi-head attention layer.
This was done to ensure that only the past and present words are available to the decoder, and the future words are masked.
The architecture allowed the decoder to have complete access to the input, instead of just considering a single vector.
The system was trained using the standard WMT 2014 English-German dataset containing about 4.5 million sentence pairs, and attained state-of-the-art results as measured by BLEU score.

Pre-training of Deep Bidirectional Transformers for Language Understanding (BERT) \cite{devlin2018bert} is a framework to generate pre-trained embeddings for sentences.
It also can be used to perform various NLP classification tasks using parameter fine-tuning after adding a single fully-connected layer.
% The authors argued that the current Deep Learning based language models to generate embeddings are unidirectional, and so there are challenges when we need to model sentences.
% Tasks such as attention based question answering require the architecture to attend to tokens before and after the current token, during the self-attention stage. 

The authors make a strong assertion that the embeddings generated by Deep Learning based language models involve a training process that is uni-directional.
Such kind of training does not learn the dependencies in both directions.
% Hence we will face challenges during the modeling of the sentences.   
In the self-attention stage, the architecture is required to attend to tokens preceding and succeeding the present token specifically for attention-based question answering tasks.

The system was trained based on two NLP tasks.
One of which was to predict a missing word, given a sentence.
To create such a training set, a large English language corpora were used.
A word was selected at random and removed subsequently.
The training objective was to predict the word that was missing from the sentence.
The other task was to identify the next suitable sentence for a given sentence, from a custom-developed set that consisted of 4 sentences.
The corpora used for training was the Books Corpus (800M words) \cite{google_books}  and English Wikipedia (2,500M words) \cite{english_wikipedia}.
We used BERT embeddings to model the sentences followed by a single layer neural network for our experiments.

The classification methods will be referred as CNN, Bi-LSTM, and BERT respectively, in the experiments section.

\subsection{Canonical Representation}
\label{section:method_canonical}
A QA pair in a legal deposition has its context spread across the question and answer. 
We wanted to use a form of the text that can combine the whole QA context into a canonical form of declarative sentences.
We hypothesize that a canonical form would be able to assist the classifier better than the question or the answer text.
We used the techniques from the work in \cite{chakravarty_jurix} to transform the QA pair into its canonical form.
We used the canonical form of a QA pair in our experiments in addition to the question and answer text.
Table \ref{table:qa_example} shows an example declarative sentence for a QA pair.

\begin{table}[h!]
\caption{A QA pair with its canonical form.}
\centering
\begin{tabular}{|p{.09\textwidth}| p{.33\textwidth}|}
 \hline
 \textbf{Type} & \textbf{Text}\\
 \hline
Question & Were you able to do physical exercises before the accident? \\ \hline
Answer & Yes. I used to play tennis before. Now I cannot stand for more than 5 minutes.\\ \hline
Canonical Form  & I was able to do physical exercises before the accident. I used to play tennis before. Now I cannot stand for more than 5 minutes. \\ \hline
\end{tabular}

\label{table:qa_example}
\end{table}

\section{Dataset}
\label{sec:Dataset}

Our classification experiments were performed using a proprietary dataset, provided by Mayfair Group LLC. 
This dataset contained accident and injury case depositions that were made available to us as a courtesy for academic purposes. 
This dataset contained about 350 depositions. 
We selected 11 depositions %from this dataset 
that pertained to 2 different litigation matters, each with multiple deponent types.
We processed the dataset for any noise and removed all of the content from the depositions other than the QA pairs. 
We ended up with a total of 9247 QA pairs that we used for training, validation, and testing. 
For our classification experiments, we divided the dataset into training, validation, and test sets into a ratio of 70, 20, and 10, respectively. 
The dataset was manually annotated by the authors. 
Table \ref{table:class_distribution} shows the class distribution for the dataset.

% \footnotesize
%\begin{tabular}{|c| c| c| c| c|}

\begin{table}[h!]
\centering
\caption{Class distribution for the dataset}
\begin{tabular}{|p{.08\textwidth}| p{.08\textwidth}| p{.08\textwidth}|}
 \hline
 \textbf{Class} & \textbf{Counts} & \textbf{\% of Total} \\
  [0.5ex]
 \hline
 B & 1455 & 15.73\\ \hline
 EB & 1468 & 15.87\\ \hline
 ED & 522 & 5.64\\ \hline
 EC & 220 & 2.38\\ \hline
 PPC & 39 & 0.42\\ \hline
 TR & 245 & 2.65\\ \hline
 EE & 62 & 0.67\\ \hline
 IP & 51 & 0.55\\ \hline
 DP & 80 & 0.86\\ \hline
 OPS & 1011 & 10.93\\ \hline
 PRD & 1617 & 17.48\\ \hline
 O & 2477 & 26.78\\ \hline
\textbf{Total} & 9247 & 100\\ \hline
\end{tabular}

\label{table:class_distribution}
\end{table}

\section{Experiment Setup and Results}
\label{sec:Experiments}

\subsection{Experimental Setup}

We wanted to understand what content from the QA pairs would enable a classification system to capture the semantics of the conversation QA pair effectively.
Is it the question, the answer, or the declarative sentence that works best?
Or would a combination of these lead to the highest achievable classification results?
We used the following combinations of the text content from the QA pairs for our experiments.

\begin{enumerate}
\item Question (Q)
\item Answer (A)
\item Question + Answer (Q + A)
\item Declarative Sentences (DS-M)
\item Question + Answer + Declarative Sentences (Q+A+DS-M)
\end{enumerate}

The declarative sentences were generated automatically by the transformation method, as described in Section \ref{section:method_canonical}. 
For the testing phase involving the DS-M input, the inference was performed using the declarative sentences as generated by the automated methods.

\subsubsection{Environment setup}
All of the classification experiments were run on a Dell server running Ubuntu 16.04,  with 32 GB RAM and two Tesla P40 NVIDIA GPUs.

\subsubsection{CNN classifier}
The implementation for this classifier was based on \cite{cnn_kimconvolutional}.
Parameters that were fine-tuned for the CNN with word2vec embeddings classifier are:
\begin{enumerate}
    \item hidden layer size: Values ranged from 100 to 500 in increments of 100.
    \item dropout: Values ranged from 0.1 to 0.5 in increments of 0.1.
    \item output layer activation functions: sigmoid, tanh, and relu.
    \item n-gram: window size based on unigram, bi-gram, and tri-gram.
    \item max-sequence length: It was kept constant at 128.
    \item batch-size: It was kept constant at 100.
    \item number of epochs: We kept it constant at 30 and performed early stopping if the validation F1-score started to drop.
\end{enumerate}

\subsubsection{LSTM classifier}
The implementation for this classifier was based on \cite{attention_bahdanau2014neural}.
Parameters that were fine-tuned for the Bi-directional LSTM with attention classifier are:
\begin{enumerate}
    \item hidden layer size: Values chosen from the set of 64, 128, and 256.
    \item embedding size: Values chosen from the set of 32, 64, 128, and 256.
    \item learning rate: Values chosen from the set of 0.0001, 0.001, 0.01, and 0.1.
    \item max-sequence length: Values chosen from the set of 32 and 128.
    \item batch-size: It was kept constant at 100.
    \item number of epochs: We kept it constant at 30 and performed early stopping if the validation F1-score started to drop.
\end{enumerate}

\subsubsection{BERT classifier}
The implementation for this classifier was based on \cite{devlin2018bert}.
Parameters that were fine-tuned for the BERT single sentence classifier are:
\begin{enumerate}
    \item learning rate: Values chosen from the set of 0.00005, 0.00002, 0.0001, 0.0005, 0.0002, 0.001, 0.005, 0.01, 0.05, and 0.1.
    \item max-sequence length: Values chosen from the set of 32 and 128.
    \item batch-size: It was kept constant at 80.
    \item number of epochs: We kept it constant at 30 and performed early stopping if the validation F1-score started to drop.
\end{enumerate}

\subsection{Results and Discussion}

Table \ref{tab:results} shows the results for each of the 3 systems. 
We attained the best F1 score of 0.83 for the BERT embeddings based classifier. 
The classifiers based on CNN and bidirectional LSTM with attention had poor performance with all of the input types.
The BERT classifier consistently outperformed all of the other systems for all input types.
For the BERT classifier, the best score was attained for the declarative sentences input.
It was surprising to us that concatenation of question and answer text had an inferior performance when used for classification, as compared to the declarative sentences.
Equally surprising was the fact that concatenating the question and answer text with the declarative sentences had a slightly lower classification performance, as compared to using just the declarative sentences.
We believe that transforming a QA pair into a declarative sentence and using that in downstream tasks would be useful because of its form and content.
The results also highlight the superior classification efficacy of the BERT based system.
The sentence embeddings generated by it were semantically rich.
Using them with a single layer neural network resulted in the best classification results, as compared to other systems.

Tables \ref{table:cnn_parameters}, \ref{table:lstm_parameters} and \ref{table:bert_parameters} list the different parameter values for the CNN, Bi-LSTM, and BERT classifiers, respectively, that yielded the best performance for the given form of text. 

% \begin{table}[h!]
% \centering
% \begin{tabular}{|p{.09\textwidth}| p{.03\textwidth}| p{.03\textwidth}| p{.03\textwidth}| p{.03\textwidth}| p{.03\textwidth}| p{.03\textwidth}| p{.03\textwidth}|}
%  \hline
%  \multirow{2}{*}{\textbf{Parameters}} & \multicolumn{7}{c|}{\textbf{Values}} \\ \cline{2-8} 
%  & \textbf{Q} & \textbf{A} & \textbf{Q+A} & \textbf{DT} & \textbf{Q+A +DT} & \textbf{DT-N} & \textbf{Q+A +DT-N}\\ \hline
%  hidden layer size & 200 & 200 & 25 & 100 & 25 & 25 & 25\\ \hline
%  dropout & 0.5 & 0.5 & 0.5 & 0.2 & 0.5 & 0.5 & 0.5\\ \hline
%  activation function & sig- moid & sig- moid & sig- moid & sig- moid & sig- moid & sig- moid & sig- moid\\ \hline
%  n-grams & tri- gram & tri- gram & tri- gram & uni- gram & tri- gram & tri- gram & tri- gram\\ \hline
% \end{tabular}
% \caption{Parameters for the CNN classifier for Deposition dataset after tuning for best performance.}
% \label{table:cnn_parameters}
% \end{table}

\begin{table}[h!]
\centering
\caption{Parameters for the CNN classifier for the deposition dataset after tuning for best performance.}
\begin{tabular}{|p{.12\textwidth}| p{.044\textwidth}| p{.044\textwidth}| p{.044\textwidth}| p{.046\textwidth}| p{.046\textwidth}|}
 \hline
 \multirow{2}{*}{\textbf{Parameters}} & \multicolumn{5}{c|}{\textbf{Values}} \\ \cline{2-6} 
 & \textbf{Q} & \textbf{A} & \textbf{Q+A} & \textbf{DS-M} & \textbf{Q+A+ DS-M}\\ \hline
 hidden layer size & 300 & 200 & 200 & 300 & 300 \\ \hline
 dropout & 0.2 & 0.2 & 0.3 & 0.5 & 0.3 \\ \hline
 activation function & sig- moid & sig- moid & sig- moid & sig- moid & sig- moid \\ \hline
 n-grams & tri- gram & uni- gram & uni- gram & uni- gram & bi- gram \\ \hline
\end{tabular}

\label{table:cnn_parameters}
\end{table}

% \begin{table}[h!]
% \centering
% \begin{tabular}{|p{.09\textwidth}| p{.03\textwidth}| p{.03\textwidth}| p{.03\textwidth}| p{.03\textwidth}| p{.03\textwidth}| p{.03\textwidth}| p{.03\textwidth}|}
%  \hline
%  \multirow{2}{*}{\textbf{Parameters}} & \multicolumn{7}{c|}{\textbf{Values}} \\ \cline{2-8} 
%  & \textbf{Q} & \textbf{A} & \textbf{Q+A} & \textbf{DT} & \textbf{Q+A +DT} & \textbf{DT-N} & \textbf{Q+A +DT-N}\\ \hline
%  hidden layer size & 128 & 128 & 64 & 256 & 64 & 256 & 64\\ \hline
%  embedding size & 128 & 64 & 256 & 256 & 128 & 128 & 128\\ \hline
%  learning rate & 0.01 & 0.01 & 0.01 & 0.01 & 0.01 & 0.01 & 0.01\\ \hline
%  max-seq-length & 32 & 32 & 32 & 32 & 128 & 32 & 128\\ \hline
%  batch-size & 100 & 100 & 100 & 100 & 100 & 100 & 100\\ \hline
%  number of epochs & 30 & 30 & 30 & 30 & 30 & 30 & 30\\ \hline
% \end{tabular}
% \caption{Parameters for the Bi-LSTM classifier for Deposition dataset after tuning for best performance.}
% \label{table:lstm_parameters}
% \end{table}

\begin{table}[h!]
\centering
\caption{Parameters for the Bi-LSTM classifier for the depo-sition dataset after tuning for best performance.}
\begin{tabular}{|p{.13\textwidth}| p{.044\textwidth}| p{.043\textwidth}| p{.044\textwidth}| p{.046\textwidth}| p{.046\textwidth}|}
 \hline
 \multirow{2}{*}{\textbf{Parameters}} & \multicolumn{5}{c|}{\textbf{Values}} \\ \cline{2-6} 
 & \textbf{Q} & \textbf{A} & \textbf{Q+A} & \textbf{DS-M} & \textbf{Q+A+ DS-M}\\ \hline
 hidden layer size & 128 & 128 & 64 & 256 & 64\\ \hline
 embedding size & 128 & 64 & 256 & 128 & 128\\ \hline
 learning rate & 0.01 & 0.01 & 0.01 & 0.01 & 0.01\\ \hline
 max-seq-length & 32 & 32 & 32 & 32 & 128\\ \hline
 batch-size & 100 & 100 & 100 & 100 & 100 \\ \hline
 number of epochs & 30 & 30 & 30 & 30 & 30 \\ \hline
\end{tabular}

\label{table:lstm_parameters}
\end{table}

% \begin{table}[h!]
% \centering
% \begin{tabular}{|p{.09\textwidth}| p{.03\textwidth}| p{.03\textwidth}| p{.03\textwidth}| p{.03\textwidth}| p{.03\textwidth}| p{.03\textwidth}| p{.03\textwidth}|}
%  \hline
%  \multirow{2}{*}{\textbf{Parameters}} & \multicolumn{7}{c|}{\textbf{Values}} \\ \cline{2-8} 
%  & \textbf{Q} & \textbf{A} & \textbf{Q+A} & \textbf{DT} & \textbf{Q+A +DT} & \textbf{DT-N} & \textbf{Q+A +DT-N}\\ \hline
%  learning rate & 2e-5 & 2e-5 & 2e-5 & 2e-5 & 2e-5 & 2e-5 & 2e-5\\ \hline
%  max-seq-length & 32 & 32 & 32 & 32 & 128 & 32 & 128\\ \hline
%  batch-size & 32 & 64 & 80 & 80 & 80 & 80 & 80\\ \hline
%  number of epochs & 30 & 30 & 30 & 30 & 30 & 30 & 30\\ \hline
% \end{tabular}
% \caption{Parameters for BERT classifier for Deposition dataset after tuning for best performance.}
% \label{table:bert_parameters}
% \end{table}

\begin{table}[h!]
\centering
\caption{Parameters for BERT classifier for the deposition dataset after tuning for best performance.}
\begin{tabular}{|p{.13\textwidth}| p{.044\textwidth}| p{.044\textwidth}| p{.044\textwidth}| p{.046\textwidth}| p{.046\textwidth}|}
 \hline
 \multirow{2}{*}{\textbf{Parameters}} & \multicolumn{5}{c|}{\textbf{Values}} \\ \cline{2-6} 
 & \textbf{Q} & \textbf{A} & \textbf{Q+A} & \textbf{DS-M} & \textbf{Q+A+ DS-M}\\ \hline
 learning rate & 2e-5 & 2e-5 & 2e-5 & 2e-5 & 2e-5\\ \hline
 max-seq-length & 32 & 32 & 32 & 32 & 128\\ \hline
 batch-size & 80 & 80 & 80 & 80 & 80\\ \hline
 number of epochs & 30 & 30 & 30 & 30 & 30\\ \hline
\end{tabular}

\label{table:bert_parameters}
\end{table}

% \begin{table}[htbp]
% \begin{tabular}{|l|l|l|l|}
% \hline
%  & CNN & Bi-LSTM & BERT \\ \hline
% Question (Q) & 0.0 & 0.4279 & 0.6398 \\ \hline
% Answer (A)& 0.0 & 0.3633 & 0.7077 \\ \hline
% Question + Answer (Q+A) & 0.0 & 0.4877 & 0.7522 \\ \hline
% Declarative Text (DT) & 0.44 & 0.5078 &  \textbf{0.9527} \\ \hline
% \begin{tabular}[c]{@{}l@{}}Question + Answer\\ + Declarative Text (Q+A+DT)\end{tabular} & 0.0 & 0.5154 & 0.8636 \\ \hline
% \begin{tabular}[c]{@{}l@{}}Declarative Text\\ 
% - Noise (DT-N)\end{tabular} & 0.0 & 0.4414 & 0.8196 \\ \hline
% \begin{tabular}[c]{@{}l@{}}Question + Answer\\ + Declarative Text - Noise \\ (Q+A+DT-N)\end{tabular} & 0.0 & 0.4868 & 0.8144 \\ \hline
% \end{tabular}
%   \caption{F1 scores for the different methods. Best score in bold.}
%   \label{tab:results}
% \end{table}

\begin{table}[htbp]
\caption{F1-scores for the different methods. Best score in bold.}
\begin{tabular}{|l|l|l|l|}
\hline
 & CNN & Bi-LSTM & BERT \\ \hline
Question (Q) & 0.44 & 0.43 & 0.64 \\ \hline
Answer (A)& 0.35 & 0.36 & 0.71 \\ \hline
Question + Answer (Q+A) & 0.47 & 0.49 & 0.75 \\ \hline
\begin{tabular}[c]{@{}l@{}}Declarative Sentences\\ 
- Machine (DS-M)\end{tabular} & 0.46 & 0.46 & \textbf{0.83} \\ \hline
\begin{tabular}[c]{@{}l@{}}Question + Answer\\ + Declarative Sentences - Machine \\ (Q+A+DS-M)\end{tabular} & 0.50 & 0.49 & 0.81 \\ \hline
\end{tabular}
  
  \label{tab:results}
\end{table}

To further explore the effectiveness of the declarative sentences, we performed another experiment.
In our analysis of the declarative sentences generated by the method in \cite{chakravarty_jurix}, we observed that a few of the generated sentences had some noise.
This was due to the less than ideal fusion of the question and answer text, which was also highlighted by the authors of the work for some QA pairs.
We wanted to explore whether using declarative sentences that are devoid of any noise would provide any improvement in the classification accuracy.
To perform this experiment, we used the same dataset and had human annotators write the declarative sentences for the QA pairs.
These annotators had learned English as their first language and were proficient in their writing ability.

We trained the BERT classifier as it was the one with the best performance.
We wanted to evaluate whether training using declarative sentences written by human annotators will improve the classification on the machine-generated declarative sentences.
We performed the classification on the test set that contained machine-generated declarative sentences.
The training was performed using the following different texts.

\begin{itemize}
  \item Machine-generated declarative sentences (DS-M)
  \item Human-written declarative sentences (DS-C)
  \item Concatenated human-written and machine-generated declarative sentences (DS-CM)
\end{itemize}

Table \ref{table:additional_training} shows the results of the experiment. 
We observed that training using the additional declarative sentences written by human annotators did not make any difference, as the F1 score remained constant at 0.83. 
This highlights that the classifier is resilient to noise, and can perform as well with noisy text as with the human-annotated sentences. 
Table \ref{table:bert_scores} lists the Precision, Recall, and F1-score for the respective classes in each of the three scenarios. 
The classification model DS-CM performs well on classes with low amounts of training data, as outlined in Table \ref{table:class_distribution}, such as EC, PPC, TR, EE, IP, and DP. 
When compared with the other two models, DS-CM was relatively better, but a test of significance yielded a p-value of .14, so the difference was not significant.
On the other hand, both the DS-M and DS-C models perform well in classes with large amounts of training data, and the DS-CM model is not far behind.

\begin{table}[h!]
\caption{Experiment results with additional training data using the BERT classifier.}
\begin{tabular}{|c|c|}
\hline
\textbf{Text used} & \textbf{F1-score} \\ \hline
DS-M & 0.83\\ \hline
DS-C & 0.82 \\ \hline
DS-CM & 0.83 \\ \hline
\end{tabular}
\label{table:additional_training}
\end{table}

The unequal distribution in the dataset is a point of concern, as observed earlier, as is how these low training data aspects affect the score of the DS-M and DS-C classification models by bringing down their weighted average.
We plan to address such in our future work.
% However, there is an increase in scores for a few of these aspects like TR and IP, when compared with our previous evaluation results, after the implementation of aspects consolidation which is covered earlier in \ref{sec:Methods-Ontology}.  

To understand the limitations of the classifier further, we performed another experiment using training and testing on human written declarative sentences.
We wanted to explore whether the classifier performs well when it has to perform inference on declarative sentences that are devoid of noise.
We trained the BERT embeddings based classifier on the dataset with human-written declarative sentences and achieved an F1-score of 0.95 on the test set.
This result shows that the presence of noise in the machine-generated declarative sentences hurts the classification efficacy.
We find this result very encouraging and believe that a detailed analysis of the classification results between the human-written and machine-generated declarative sentences would provide some insights in tuning the classifiers further.
We plan to address this in our future work.

\begin{table}[h!]
\centering
\caption{Classification scores for BERT}
\begin{tabular}{|p{.03\textwidth}| p{.03\textwidth}| p{.03\textwidth}| p{.03\textwidth}| p{.03\textwidth}| p{.03\textwidth}|p{.03\textwidth}|p{.03\textwidth}|p{.03\textwidth}|p{.03\textwidth}|}
 \hline
 \multirow{2}{*}{\textbf{Cls}} & \multicolumn{3}{c|}{\textbf{DS-M}}  & \multicolumn{3}{c|}{\textbf{DS-C}}  & \multicolumn{3}{c|}{\textbf{DS-CM}}\\ \cline{2-10} 
 & \textbf{P} & \textbf{R} & \textbf{F1} & \textbf{P} & \textbf{R} & \textbf{F1} & \textbf{P} & \textbf{R} & \textbf{F1}\\ \hline
 B & 0.92 & 0.91 & \textbf{0.91} & 0.89 & 0.89 & 0.89 & 0.89 & 0.90 & 0.90\\ \hline
 EB & 0.86 & 0.86 & 0.86 & 0.89 & 0.84 & \textbf{0.87} & 0.79 & 0.87 & 0.83\\ \hline
 ED & 0.84 & 0.75 & 0.79 & 0.88 & 0.86 & \textbf{0.87} & 0.89 & 0.81 & 0.85\\ \hline
 EC & 0.74 & 0.81 & 0.77 & 0.55 & 0.69 & 0.61 & 0.80 & 0.89 & \textbf{0.84}\\ \hline
 PPC & 1.00 & 1.00 & \textbf{1.00} & 1.00 & 1.00 & \textbf{1.00} & 1.00 & 1.00 & \textbf{1.00}\\ \hline
 TR & 0.59 & 0.76 & 0.67 & 0.71 & 0.73 & 0.72 & 0.75 & 0.92 & \textbf{0.83}\\ \hline
 EE & 0.83 & 1.00 & 0.91 & 0.71 & 1.00 & 0.83 & 1.00 & 1.00 & \textbf{1.00}\\ \hline
 IP & 0.80 & 0.80 & 0.80 & 0.43 & 0.60 & 0.50 & 0.78 & 1.00 & \textbf{0.88}\\ \hline
 DP & 0.57 & 0.57 & 0.57 & 0.50 & 0.71 & 0.59 & 0.86 & 0.80 & \textbf{0.83}\\ \hline
 OPS & 0.77 & 0.79 & 0.78 & 0.75 & 0.84 & \textbf{0.79} & 0.71 & 0.75 & 0.73\\ \hline
 PRD & 0.81 & 0.89 & \textbf{0.85} & 0.83 & 0.88 & \textbf{0.85} & 0.82 & 0.85 & 0.84\\ \hline
 O & 0.84 & 0.76 & \textbf{0.80} & 0.82 & 0.71 & 0.76 & 0.87 & 0.73 & 0.79\\ \hline \hline
 Avg. & 0.83 & 0.83 & \textbf{0.83} & 0.83 & 0.82 & 0.82 & 0.83 & 0.83 & \textbf{0.83}\\ \hline
\end{tabular}

\label{table:bert_scores}
\end{table}

\subsection{Error Analysis}
We chose the best performing classification system results, as shown in Table \ref{tab:results}, and performed a detailed error analysis on the misclassifications.
For the analysis, we only included classes that had either an F1-score < 0.9, or a number of misclassifications greater than a threshold of 10.
Table \ref{table:error-analysis} discusses the errors associated with each aspect.

\begin{table*}[h!]
\caption{Error analysis}
\begin{tabular}{|p{.22\linewidth}|p{.75\linewidth}|}
\hline
\multicolumn{1}{|c|}{\textbf{Class}} & \multicolumn{1}{c|}{\textbf{Analysis}} \\ \hline
B (Biographical) & Most of the misclassifications are attributed to the incorrect class assignment to ``OPS," ``O," and ``EB" classes. Sentences assigned to ``OPS" and ``EB" categories have similar words often found in the listing of skills of a person, which are used in the ``B" category such as ``work," ``manufacturing," ``construction." This inclusion of these words without proper context results in misclassification. The declarative sentences generated by the automated methods sometimes remove the context. For example, a human-written declarative sentence such as ``i left ORG165 due to difference of opinion" includes the background with not so much reliance on other sentences in the block. But in a machine-generated declarative sentence, ``i did leave there due to difference of opinion" does not convey the context unless the previous and preceding declarative sentences are also included. Such sentences are often assigned to the ``O" class.\\ \hline
EB (Event Background) & With the exception of getting misclassified to ``B" and ``OPS" categories for the similar reasons mentioned in the ``B" category, most get assigned to ``ED". This misclassification has to do with the sharing of certain words in the training example with the ``ED" and ``EC" classes.\\ \hline
ED (Event Details) & The few misclassifications for this class are assigned to the ``O" class. Mostly these sentences do not include the essential terms/context but are just repeated sentences by a deponent during the deposition.  \\ \hline
EC (Event Consequences) & Training data for this class is less, making it hard for the classifier to learn words associated with this class as compared to ``ED," ``EB," ``IP". \\ \hline
% PPC (Prior Physical Condition) & Almost negligible incidents of misclassification of class. \\ \hline
TR (Treatments Received) & The misclassifications can be attributed to the lack of context in the sentences. The sharing of terms in the training data with the ``PPC" class makes learning the correct word associations difficult for the classifier.\\ \hline
% EE (Expert Elaboration)    & Numerous unique words associated with this class made it less prone to misclassification even with low training data. \\ \hline
IP (Impact on Plaintiff) & The medical impact of the injury on the plaintiff as against the other kind of impacts is often misclassified in this class. The misclassified sentences are assigned to the ``EE" class. The inclusion of terminology in these classes is the primary reason for this misclassification. \\ \hline
DP (Deposition Procedures)  & Possible misclassifications in this class are due to a limited amount of training data and frequent use of the word ``deposition," which is also found in the ``B" class when enquiring about past depositions of the deponent. \\ \hline
OPS (Operational procedures/ inspections/ maintenance/ repairs) & The majority of the misclassifications are assigned to ``EB" and ``PRD" classes. A common reason for getting attached to an event-related class is the terminology cross over. ``OPS" and ``PRD" classes are only covered as part of the witness deponent roles such as Related Organization Witness, leading to the sharing of words in the training data.\\ \hline
PRD (Plaintiff-related Details)     & Most misclassifications can be attributed to the ``B" class. Conversations about acquaintance with the plaintiff and other related people are mostly considered as part of the ``B" class. The presence of the word ``plaintiff" sometimes results in misclassification. \\ \hline
O (Other) & Sentences not assigned to any of the previous classes are all assigned to ``O", making it a mix of many variants of terminology and context. The sharing of terms with other classes results in the assignment of the sentences of this class to other classes, leading to low recall.\\ \hline
\end{tabular}

\label{table:error-analysis}
\end{table*}

\subsection{Limitations/Threat to Validation}
We want to highlight the following limitations of our work. We plan to address some of them in our future work and continued research in this area.
\begin{itemize}
    \item We used a total of 11 depositions from 2 different cases. The small scale of this dataset might not be representative of all the aspects in accident and injury-related cases.
    \item We had a total of 9247 QA pairs used for our classification work. Expanding this to a larger size would increase the performance of our classifier, but owing to the high cost of annotation of such data, we were limited in the scale of data used in our experiments.
    \item The Deep Learning landscape is changing very fast in recent times. Text classification is continually being improved by researchers in the field. Our methods might not be the best methods in text classification at the time of publication.
\end{itemize}

\section{Conclusion and Future Work}
\label{sec:Conclusion}
% State-of-the-art summarization methods and traditional NLP techniques are challenging to apply to question-answer pairs.
% Our preliminary testing with summarization methods applied to QA pairs led to poor results.
Traditional and state-of-the-art NLP and summarization techniques are challenging to apply to corpora that consist of question-answer pairs.
Applying summarization methods as-is on QA pairs leads to less than ideal results because of their form.
We used a transformation of the QA pair to a canonical form to ease the processing of such text for classification.
However, using the canonical form did not improve the summarization results by much.
Hence we desire a summarization method that is customized for the legal domain, but can be extended to other domains.
Having a way to break a legal deposition into its constituent aspects topically would help segment the various parts of the depositions that would be further useful for summarization.
To achieve this, we developed an ontology of aspects that pertains to the legal domain, especially for property and casualty insurance claims. 
This ontology can be expanded or modified for other kinds of cases or for different domains.
For classification purposes, we have developed our own dataset by manually annotating 9247 QA pairs for their respective aspects.
This dataset helped us in training and evaluating our classifiers.

We developed three classification methods for classifying the aspects present in legal depositions:
\begin{itemize}
\item Convolutional Neural Network (CNN) with word2vec embeddings,
\item Bi-directional Long Short Term Memory (LSTM) with attention mechanism, and
\item BERT sentence embeddings based classifier.
\end{itemize}

We plan to improve and extend this work in the following ways.
\begin{enumerate}
\item Add more training examples to the dataset for the aspects that had classification errors due to a smaller number of training examples.
\item Provide more context to the classifier using the preceding classes from the previous 2 QA pairs \cite{context_da} to classify the current QA pair, to improve the classification accuracy.
\item Not all depositions will have all the aspects. Creating a framework to group certain aspects into specific ones based on deponent type would further help organize aspects better.
\item Analyze the classifications on the test set involving human written declarative sentences. Identify the classification differences vis-a-vis with the machine-generated declarative sentences to improve the classifiers further.
\item We have collected additional data for classification. Attorneys and paralegals have annotated this data. Training the system on this additional data should increase the classification score further.
\item Develop NLP and deep learning techniques to identify segments within the legal deposition that are centered around the same topic. Using the aspects would help create these segments that can be used to generate summaries.
% \item Use learning to rank \cite{learntorank_liu2009learning} to order the sentences for the aspects based on their relative importance as per the case facts. These can be used as candidate sentences to select to form a summary.
% \item Identify the statistics distribution and aspect based layout of different aspects in a deposition summary and use a linear optimization method to select declarative sentences from the ordered sentences based on aspects to generate a summary that follows the same distribution and layout.
\item Develop explainable AI methods to correlate summary sentences back to the parts in the deposition that provide their support.
%where they were sourced from.
\end{enumerate}

\section*{Acknowledgments}
This work was made possible by the Virginia Tech's Digital Library Research Laboratory (DLRL). 
Data in the form of legal depositions was provided by Mayfair Group LLC.
In accordance with Virginia Tech policies and procedures and my ethical obligation as a researcher, we are reporting that Dr. Edward Fox has an equity interest in Mayfair Group, LLC. whose data was used in this research. Dr. Fox has disclosed those interests fully to Virginia Tech, and has in place an approved plan for managing any potential conflicts arising from this relationship.

\bibliographystyle{unsrt}
\bibliography{sample-base}

\begin{thebibliography}{10}

\bibitem{chakravarty_jurix}
Saurabh Chakravarty, Maanav Mehrotra, Raja Venkata Satya~Phanindra Chava, Han
  Liu, Matthew Krivansky, and Edward~A. Fox.
\newblock Improving the processing of question answer based legal documents.
\newblock In {\em Proceedings of the JURIX 2019 Conference: Legal Knowledge and
  Information Systems}, 2019.

\bibitem{DA}
Daniel Jurafsky, Elizabeth Shriberg, Barbara Fox, and Traci Curl.
\newblock Lexical, prosodic, and syntactic cues for dialog acts.
\newblock {\em Journal on Discourse Relations and Discourse Markers}, 1998.

\bibitem{da_classic1}
Jeremy Ang, Yang Liu, and Elizabeth Shriberg.
\newblock Automatic dialog act segmentation and classification in multiparty
  meetings.
\newblock In {\em Proceedings ICASSP'05. IEEE International Conference on
  Acoustics, Speech, and Signal Processing, 2005.}, volume~1, pages I--1061.
  IEEE, 2005.

\bibitem{da_classic2}
Gang Ji and Jeff Bilmes.
\newblock Dialog act tagging using graphical models.
\newblock In {\em Proceedings ICASSP'05. IEEE International Conference on
  Acoustics, Speech, and Signal Processing, 2005.}, volume~1, pages I--33.
  IEEE, 2005.

\bibitem{da_classic3}
Anand Venkataraman, Luciana Ferrer, Andreas Stolcke, and Elizabeth Shriberg.
\newblock Training a prosody-based dialog act tagger from unlabeled data.
\newblock In {\em 2003 IEEE International Conference on Acoustics, Speech, and
  Signal Processing, 2003. Proceedings.(ICASSP'03).}, volume~1, pages I--I.
  IEEE, 2003.

\bibitem{da_classic4}
N~Webb, M~Hepple, and Y~Wilks.
\newblock Dialog act classification based on intra-utterance features.
  cs-05-01.
\newblock {\em Dept. of Computer Science, University of Sheffield, UK}, 2005.

\bibitem{da_classic5}
Raul Fernandez and Rosalind~W Picard.
\newblock Dialog act classification from prosodic features using support vector
  machines.
\newblock In {\em Speech Prosody 2002, International Conference}, 2002.

\bibitem{da_classic6}
Yang Liu.
\newblock Using {SVM} and error-correcting codes for multiclass dialog act
  classification in meeting corpus.
\newblock In {\em Ninth International Conference on Spoken Language
  Processing}, 2006.

\bibitem{da_classic7}
Marion Mast, Ralf Kompe, Stefan Harbeck, Andreas Kie{\ss}ling, Heinrich
  Niemann, Elmar Noth, Ernst~G{\"u}nter Schukat-Talamazzini, and Volker Warnke.
\newblock Dialog act classification with the help of prosody.
\newblock In {\em Proceeding of Fourth International Conference on Spoken
  Language Processing. ICSLP'96}, volume~3, pages 1732--1735. IEEE, 1996.

\bibitem{da_classic8}
Pavel Kr{\'a}l and Christophe Cerisara.
\newblock Automatic dialogue act recognition with syntactic features.
\newblock {\em Language resources and evaluation}, 48(3):419--441, 2014.

\bibitem{tobacco-link}
{UCSF Library}.
\newblock {\em Truth Tobacco Industry Documents}, 2002.
\newblock \url{https://www.industrydocuments.ucsf.edu/tobacco}.

\bibitem{da_for_qa}
Saurabh Chakravarty, Raja Venkata Satya~Phanindra Chava, and Edward~A. Fox.
\newblock Dialog acts classification for question-answer corpora.
\newblock In {\em Proceedings of the Third Workshop on Automated Semantic
  Analysis of Information in Legal Text (ASAIL 2019), June 21, 2019, Montreal,
  QC, Canada}, 2019.

\bibitem{classical_NB}
Irina Rish et~al.
\newblock An empirical study of the naive {B}ayes classifier.
\newblock In {\em IJCAI 2001 workshop on empirical methods in artificial
  intelligence}, volume~3, pages 41--46, 2001.

\bibitem{classical_DT}
S~Rasoul Safavian and David Landgrebe.
\newblock A survey of decision tree classifier methodology.
\newblock {\em IEEE transactions on systems, man, and cybernetics},
  21(3):660--674, 1991.

\bibitem{classical_nearest_neighbor}
Thomas Cover and Peter Hart.
\newblock Nearest neighbor pattern classification.
\newblock {\em IEEE transactions on information theory}, 13(1):21--27, 1967.

\bibitem{classical_association_rules}
Rakesh Agrawal, Ramakrishnan Srikant, et~al.
\newblock Fast algorithms for mining association rules.
\newblock In {\em Proc. 20th int. conf. very large data bases, VLDB}, volume
  1215, pages 487--499, 1994.

\bibitem{fs_gini}
Elchanan~Ben Porath and Itzhak Gilboa.
\newblock Linear measures, the gini index, and the income-equality trade-off.
\newblock {\em Journal of Economic Theory}, 64(2):443--467, 1994.

\bibitem{fs_conditional_entropy}
Alberto Porta, Stefano Guzzetti, Nicola Montano, Raffaello Furlan, Massimo
  Pagani, Alberto Malliani, and Sergio Cerutti.
\newblock Entropy, entropy rate, and pattern classification as tools to typify
  complexity in short heart period variability series.
\newblock {\em IEEE Transactions on Biomedical Engineering}, 48(11):1282--1291,
  2001.

\bibitem{fs_chi_square}
Janez Brank, Marko Grobelnik, Natasa Milic-Frayling, and Dunja Mladenic.
\newblock Interaction of feature selection methods and linear classification
  models.
\newblock In {\em Workshop on Text Learning held at ICML}, 2002.

\bibitem{fs_pmi}
M~Ikonomakis, Sotiris Kotsiantis, and V~Tampakas.
\newblock Text classification using machine learning techniques.
\newblock {\em WSEAS transactions on computers}, 4(8):966--974, 2005.

\bibitem{glove}
Jeffrey Pennington, Richard Socher, and Christopher Manning.
\newblock Glove: Global vectors for word representation.
\newblock In {\em Proceedings of the 2014 conference on empirical methods in
  natural language processing (EMNLP)}, pages 1532--1543, 2014.

\bibitem{word2vec}
Tomas Mikolov, Ilya Sutskever, Kai Chen, Greg~S Corrado, and Jeff Dean.
\newblock Distributed representations of words and phrases and their
  compositionality.
\newblock In {\em Advances in neural information processing systems}, pages
  3111--3119, 2013.

\bibitem{iyyer2015deep}
Mohit Iyyer, Varun Manjunatha, Jordan Boyd-Graber, and Hal Daum{\'e}~III.
\newblock Deep unordered composition rivals syntactic methods for text
  classification.
\newblock In {\em Proceedings of the 53rd Annual Meeting of the Association for
  Computational Linguistics and the 7th International Joint Conference on
  Natural Language Processing (Volume 1: Long Papers)}, pages 1681--1691, 2015.

\bibitem{castro2017classifying}
Eduardo~PS Castro, Saurabh Chakravarty, Eric Williamson, Denilson~Alves
  Pereira, and Edward~A Fox.
\newblock Classifying short unstructured data using the {A}pache {S}park
  platform.
\newblock In {\em Proceedings of the 17th ACM/IEEE Joint Conference on Digital
  Libraries}, pages 129--138. IEEE Press, 2017.

\bibitem{encoder_decoder_cho2014properties}
Kyunghyun Cho, Bart van Merrienboer, Dzmitry Bahdanau, and Yoshua Bengio.
\newblock On the properties of neural machine translation: Encoder--decoder
  approaches.
\newblock In {\em Proceedings of SSST-8, Eighth Workshop on Syntax, Semantics
  and Structure in Statistical Translation}, pages 103--111, 2014.

\bibitem{attention_bahdanau2014neural}
Dzmitry Bahdanau, Kyunghyun Cho, and Yoshua Bengio.
\newblock Neural machine translation by jointly learning to align and
  translate.
\newblock In {\em Proceedings of International Conference on Learning
  Representations (ICLR)}, 2014.

\bibitem{devlin2018bert}
Jacob Devlin, Ming{-}Wei Chang, Kenton Lee, and Kristina Toutanova.
\newblock {BERT:} pre-training of deep bidirectional transformers for language
  understanding.
\newblock In Jill Burstein, Christy Doran, and Thamar Solorio, editors, {\em
  Proceedings of the 2019 Conference of the North American Chapter of the
  Association for Computational Linguistics: Human Language Technologies,
  {NAACL-HLT} 2019, Minneapolis, MN, USA, June 2-7, 2019, Volume 1 (Long and
  Short Papers)}, pages 4171--4186. Association for Computational Linguistics,
  2019.

\bibitem{vaswani2017attention}
Ashish Vaswani, Noam Shazeer, Niki Parmar, Jakob Uszkoreit, Llion Jones,
  Aidan~N Gomez, \L~ukasz Kaiser, and Illia Polosukhin.
\newblock Attention is {A}ll you {N}eed.
\newblock In I.~Guyon, U.~V. Luxburg, S.~Bengio, H.~Wallach, R.~Fergus,
  S.~Vishwanathan, and R.~Garnett, editors, {\em Advances in Neural Information
  Processing Systems 30}, pages 5998--6008. Curran Associates, Inc., 2017.

\bibitem{cnn_kimconvolutional}
Yoon Kim.
\newblock Convolutional neural networks for sentence classification.
\newblock In {\em Proceedings of the 2014 Conference on Empirical Methods in
  Natural Language Processing (EMNLP)}, pages 1746–--1751. ACL, 2014.

\bibitem{goldberg2017neural}
Yoav Goldberg.
\newblock Neural network methods for natural language processing.
\newblock {\em Synthesis Lectures on Human Language Technologies},
  10(1):1--309, 2017.

\bibitem{cnn_understanding}
Denny Britz.
\newblock Understanding convolutional neural networks for {NLP}.
\newblock {\em Accessed May 15, 2020 from URL: http://www. wildml.
  com/2015/11/understanding-convolutional-neuralnetworks-for-nlp/}, 2020.

\bibitem{lstm}
Sepp Hochreiter and J{\"u}rgen Schmidhuber.
\newblock Long short-term memory.
\newblock {\em Neural computation}, 9(8):1735--1780, 1997.

\bibitem{bi-lstm-attention}
Peng Zhou, Wei Shi, Jun Tian, Zhenyu Qi, Bingchen Li, Hongwei Hao, and Bo~Xu.
\newblock Attention-based bidirectional long short-term memory networks for
  relation classification.
\newblock In {\em Proceedings of the 54th Annual Meeting of the Association for
  Computational Linguistics (Volume 2: Short Papers)}, volume~2, pages
  207--212, 2016.

\bibitem{dropout}
Nitish Srivastava, Geoffrey Hinton, Alex Krizhevsky, Ilya Sutskever, and Ruslan
  Salakhutdinov.
\newblock Dropout: a simple way to prevent neural networks from overfitting.
\newblock {\em The Journal of Machine Learning Research}, 15(1):1929--1958,
  2014.

\bibitem{google_books}
{Mark Davies}.
\newblock Google books corpora, 2011.
\newblock [Online; accessed 28-April-2019].

\bibitem{english_wikipedia}
William Coster and David Kauchak.
\newblock Simple {E}nglish {W}ikipedia: a new text simplification task.
\newblock In {\em Proceedings of the 49th Annual Meeting of the Association for
  Computational Linguistics: Human Language Technologies: short papers-Volume
  2}, pages 665--669. Association for Computational Linguistics, 2011.

\bibitem{context_da}
Chandrakant Bothe, Cornelius Weber, Sven Magg, and Stefan Wermter.
\newblock A context-based approach for dialogue act recognition using simple
  recurrent neural networks.
\newblock In {\em Proceedings of the Eleventh International Conference on
  Language Resources and Evaluation (LREC-2018)}, 2018.

\end{thebibliography}
% \bibliography{sample-base}
% \bibliographystyle{acm}
\end{document}